\documentclass[lettersize,journal]{IEEEtran}
\usepackage{amsmath,amsfonts}
\usepackage{algorithmic}
\usepackage{algorithm}
\usepackage{array}
\usepackage[caption=false,font=normalsize,labelfont=sf,textfont=sf]{subfig}
\usepackage{textcomp}
\usepackage{stfloats}
\usepackage{url}
\usepackage{verbatim}
\usepackage{graphicx}
\usepackage{cite}
\begin{document}

\title{Extended Preintegration for Relative State Estimation of Leader-Follower Platform}

\author{Ruican Xia, Hailong Pei
\thanks{Ruican Xia and Hailong Pei are with the Key Laboratory of Autonomous Systems and Networked Control, Ministry of Education, Unmanned Aerial Vehicle Systems Engineering Technology Research Center of Guangdong, South China University of Technology, Guangzhou, 510640, China.}
\thanks{This work was supported in part by the Scientific Instruments Development Program of NSFC [61527810], Science and Technology Planning Project of Guangdong, China [2017B010116005] and “the Fundamental Research Funds for the Central Universities”.}}

\markboth{Journal of \LaTeX\ Class Files,~Vol.~14, No.~8, August~2021}%
{Shell \MakeLowercase{\textit{et al.}}: A Sample Article Using IEEEtran.cls for IEEE Journals}

\IEEEpubid{0000--0000/00\$00.00~\copyright~2021 IEEE}

\maketitle

\begin{abstract}
Relative state estimation using exteroceptive sensors suffers from limitations of the field of view (FOV) and false detection, that the proprioceptive sensor (IMU) data are usually engaged to compensate. Recently ego-motion constraint obtained by Inertial measurement unit (IMU) preintegration has been extensively used in simultaneous localization and mapping (SLAM) to alleviate the computation burden. This paper introduces an extended preintegration incorporating the IMU preintegration of two platforms to formulate the motion constraint of relative state. One merit of this analytic constraint is that it can be seamlessly integrated into the unified graph optimization framework to implement the relative state estimation in a high-performance real-time tracking thread, another point is a full smoother design with this precise constraint to optimize the 3D coordinate and refine the state for the refinement thread. We compare extensively in simulations the proposed algorithms with two existing approaches to confirm our outperformance. In the real virtual reality (VR) application design with the proposed estimator, we properly realize the visual tracking of the six degrees of freedom (6DoF) controller suitable for almost all scenarios, including the challenging environment with missing features, light mutation, dynamic scenes, etc. The demo video is at https://www.youtube.com/watch?v=0idb9Ls2iAM. For the benefit of the community, we make the source code public.
\end{abstract}


\section{Introduction}
Relative state estimation refers to tracking the relative pose (attitude and translation) and velocity of two platforms with applications of close formation, aviation docking, collaborative simultaneous localization and mapping, entertainment\cite{ref2}, etc. The estimation is generally realized by relative measurement from exteroceptive sensors such as the pairwise distance between ultra-wideband (UWB) devices\cite{ref2}, visual features of pre-installed marks like LED array \cite{ref42} or the Apriltags \cite{ref13} etc. The measurement is used to calculate the relative pose and initialize a relative state estimator. For the ranging measurement, Chepuri et al. \cite{ref41} derived the transformation from raw pairwise distances to the orthogonal Procrustes problem and obtained the numerical solution of the relative pose. For the visual features, the measurement is used by the perspective-n-point (PNP) solver \cite{ref43}\cite{ref44} to calculate the relative pose, where the typical methods to attain numerical solution include closed-form quaternion\cite{ref45}, TRIAD algorithm\cite{ref46} and Q-method\cite{ref47}.


Although relative measurement from exteroceptive sensors can provide relative pose in proper situations, limited FOV and detection range \cite{ref2}, non-line-of-sight (NLOS) \cite{ref49}, multipath effect \cite{ref50}, etc., can noticeably impact the measurement and further the estimation. To solve the problem, the proprioceptive IMU is used in the estimator after proper time synchronization, which can be realized with the UWB network \cite{ref48}. Fosbury et al. propose general relative kinematics to fuse IMU measurement with overlapping visual features from both platforms \cite{ref0}. The kinematics is formulated with true IMU measurement of the leader and relative acceleration composed of the Euler acceleration calculated from true applied torque or angular acceleration. Using the approximate torque or angular acceleration and biased IMU measurement with noise is equivalent to including the unmodeled process noise and decreases the robustness. Our last work presents a relative kinematic with the acceleration of both platforms instead of the relative acceleration for a filter-based estimator\cite{ref9}. However, the leader’s IMU measurement error is still ignored. 

IMU preintegration has become popular in the recent decade as a factor in unified graph optimization for SLAM \cite{ref10}. This factor enables the fusion of IMU with exteroceptive sensors to estimate the current state and refine the entire state of history with less time expanse. Prompted by \cite{ref0}\cite{ref9}\cite{ref10}, this paper proposes a factor named extended preintegration with analytic derivation to incorporate the IMU preintegration from both platforms for relative state estimation and the entire history refinement. Demonstration videos and the code can be found in our project webpage\footnote{https://github.com/richardXia7462/RelativeStateEstimation}.

This paper is organized as follows. Section \ref{sec1} details specific problem formulation and measurement models. Section \ref{sec2} proposes the extended preintegration with state and uncertainty propagation, residual definition, and associated Jacobian. Based on the extended preintegration, we propose the algorithm of the relative state estimator for the current state and the full smoother for the entire state history. Section \ref{sec3} performs the simulation to compare the proposed methods with the previous algorithms from the perspective of trajectory dynamic and measurement noise, and the real experiment to realize the visual tracking of the 6DoF controller in the VR scenario using the proposed algorithm with ground truth evaluation. Section \ref{sec4} gives the conclusion of the overall paper with prospects.

\section{System Overview}\label{sec1}
\subsection{Problem Formulation}
This paper focuses on the problem of instantaneous relative state estimation of the leader-follower platform. The state of interest includes the relative pose, velocity, and the biases of both platforms, i.e.
\begin{equation}
	X=\left\lbrace \mathrm{R}_F^L, \mathbf{t}_{F\mid L}^{L}, \dot{\mathbf{t}}_{F \mid L}^{L}, \boldsymbol{\beta}_{Fg}, \boldsymbol{\beta}_{Fa}, \boldsymbol{\beta}_{Lg}, \boldsymbol{\beta}_{La}\right\rbrace,
\end{equation}
where $L$ and $F$ are the body frames of the leader and the follower, $\mathrm{R}_{F}^{L}$ is the rotation convert vector from frame $F$ to frame $L$, $\mathbf{t}_{F\mid L}^{L}$ is the translation of the frame $F$ w.r.t. frame $L$ given in frame $L$, $\boldsymbol{\beta}_{F(L)g}$ and $\boldsymbol{\beta}_{F(L)a}$ are the gyroscope and accelerometer biases of the follower (leader). 
The collaborative platforms are each equipped with an IMU and other sensors to provide relative measurements such as the visual feature, pairwise distance, etc. The solution addresses the general 6DoF relative motion instead of the orbit application\cite{ref1}. Measurement of applied torque of the platforms used in \cite{ref0} is unavailable with the consideration of universal. The global state from odometry \cite{ref2}-\cite{ref8} or GPS is also inaccessible because the application scenario can be an indoor environment without texture.  

Attitude-involved estimation problems are nonlinear for the properties of rotation group. Both the filter-based and optimization-based estimators required appropriate linearization points, propagated from the previous posterior state with inertial measurements. 
The precise propagation is generally more significant in this problem than the inertial localization because the relative translation and velocity can be sensitive to the rotation of the reference frame depending on the distance between the platforms. Our last work provides a solution with a filter-based estimator\cite{ref9}, but the optimization-based method with full smoother is necessary for applications with high precision requirements and the algorithm is still underdeveloped.

\subsection{Measurement Model}
The IMU measurements contain linear acceleration and angular velocity corrupted with noise and biases. For both platforms, the measurement model is defined as\cite{ref11}
\begin{align}\label{eq1_}
	\tilde{\mathbf{a}}_{B} & =\mathrm{R}_{I}^{B}\left(\ddot{\mathbf{t}}_{B \mid I}^{I}-\mathbf{g}^{I}\right)+\boldsymbol{\beta}_{Ba}+\boldsymbol{\eta}_{Ba v} \\
	& =\mathbf{a}_{B}+\boldsymbol{\beta}_{Ba}+\boldsymbol{\eta}_{Ba v}, (B=F,L)\nonumber
\end{align}
\begin{equation}\label{eq2_}
	\tilde{\boldsymbol{\omega}}_{B}=\boldsymbol{\omega}_{B}+\boldsymbol{\beta}_{Bg}+\boldsymbol{\eta}_{Bg v},
\end{equation}
where $I$ is the inertial frame,
$\mathbf{g}$ is the gravity, $\left(\tilde{\cdot}\right)$ is the measurement with noise,
  $\boldsymbol{\eta}_{Ba v}$ and $\boldsymbol{\eta}_{Bg v}$ are supposed to be independent identically distributed (i.i.d.) zero-mean white random processes with variances $\sigma_{Bav}^2\mathrm{I}_{3\times3}$ and $\sigma_{Bgv}^2\mathrm{I}_{3\times3}$. 

Both biases are modeled as a random walk with derivatives:
\begin{equation}\label{eq3_}
	\dot{\boldsymbol{\beta}}_{Ba}=\boldsymbol{\eta}_{Ba u}, \quad \dot{\boldsymbol{\beta}}_{Bg}=\boldsymbol{\eta}_{Bg u} ,
\end{equation}
where $\boldsymbol{\eta}_{Ba u}$ and $\boldsymbol{\eta}_{Bg u}$ are i.i.d. zero-mean white random processes with variances $\sigma_{Bau}^2\mathrm{I}_{3\times3}$ and $\sigma_{Bgu}^2\mathrm{I}_{3\times3}$. 

The relative measurement used in this paper is visual features from specific marks. The features with known coordinates given in the body frame denoted as $\mathbf{t}_{k \mid B}^{B}(k\in{\varTheta})$, where $\varTheta$ is the set composed of all features on the platform. We fix the marks on the follower and equip the leader with a camera so the measurement model is
\begin{equation}\label{eq4_}
	\tilde{\boldsymbol{\rho}}_k = \pi\left(\mathbf{t}_{k \mid C}^{C}\right)+\boldsymbol{\eta}_{\rho k},
\end{equation}
where $\pi\left(\cdot\right)$ is a standard perspective projection based on the specific camera model, $C$ is the camera frame and
\begin{equation}\label{eq5_}
	\mathbf{t}_{k \mid C}^{C} = \mathrm{R}_L^C\left(\mathrm{R}_F^L \mathbf{t}_{k \mid F}^{F}+\mathbf{t}_{F \mid L}^{L}\right)+\mathbf{t}_{L \mid C}^{C},
\end{equation}
$\boldsymbol{\eta}_{\rho k}$ is the i.i.d. zeros-means white noise with variance $\sigma_{\rho k}^2\mathrm{I}_{2\times2}$.

\section{Extended Preintegration}\label{sec2}
\begin{figure}[!t]
	\centering
	\includegraphics[width=2.5in]{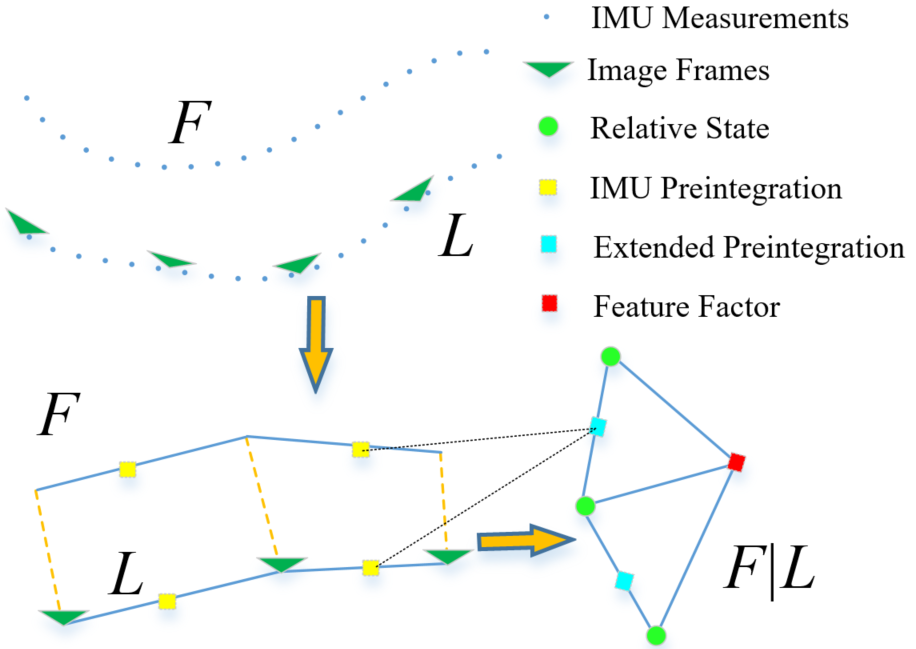}
	\caption{Relative state estimator based on the extended preintegration. Top: visual and inertial measurements sampled by platforms $L$ and $F$. Bottom: IMU preintegration is calculated and they are further converted to an extended preintegration, serves as factors to fuse with feature factors.}
	\label{fig_2}
\end{figure}
This section derives the extended preintegration of the leader-follower platform as shown in Fig.\ref{fig_2} with IMU preintegration \cite{ref10}. The extended preintegration is a factor involved in the relative pose, velocity, and biases of IMU on both platforms. We present the relative state and the uncertainties propagation, residual definition, and Jacobian calculation in an analytic form.


\subsection{Propagation of Relative State and Uncertainty}
Given attitude preintegration $\Delta {\mathrm{R}_{F(L)ij}}$ of the follower (leader) between time instances $i$ and $j$, we can formulate the relative attitude as
\begin{equation}\label{eq16}
	\mathrm{R}_{Fj}^{L}=\Delta {\mathrm{R}_{Li j}^{\top}} \mathrm{R}_{Fi}^{L} \Delta \mathrm{R}_{Fij},
\end{equation}
where $\left(\cdot\right)_{i(j)}$ is the variable at instance $i(j)$.
IMU preintegrations from both platforms use inertial measurements during the same interval where linear interpolation is performed at both ends as shown by the orange dash lines of Fig.\ref{fig_2}.

Derivation of the relative velocity and translation are less straightforward. We respectively convert the velocity and translation preintegration of the follower to frame $L$ and subtract the counterpart of the leader, i.e.
\begin{align}
	&\mathrm{R}_{Fi}^{L} \Delta \dot{\mathbf{t}}_{Fij}-\Delta \dot{\mathbf{t}}_{Li j}=\Delta \mathrm{R}_{Li j} \mathrm{R}_{Ij}^{L} \dot{\mathbf{t}}_{F \mid L j}^{I}-\mathrm{R}_{I i}^{L} \dot{\mathbf{t}}_{F \mid Li}^{I},\label{eq17}\\
	&\mathrm{R}_{Fi}^{L} \Delta \mathbf{t}_{Fi j}-\Delta \mathbf{t}_{Li j}=\Delta \mathrm{R}_{Li j} \mathbf{t}_{F \mid Lj}^{L}-\mathbf{t}_{F \mid Li}^{L}-\mathrm{R}_{Ii}^{L} \dot{\mathbf{t}}_{F \mid Li}^{I} T,\label{eq18}
\end{align}
where $T$ is the duration of the interval, $(\cdot)^{\wedge}$ converts a vector in $\mathbb{R}^{3}$ to a skew matrix. Readers can refer to \cite{ref10} for the detail of IMU preintegration. Using the velocity transformation
\begin{equation}\label{eq19}
	\mathrm{R}_{Ii}^{L} \dot{\mathbf{t}}_{F \mid Li}^{I}=\dot{\mathbf{t}}_{F \mid Li}^{L}+\boldsymbol{\omega}_{Li}^{\wedge} \mathbf{t}_{F \mid Li}^{L},
\end{equation}
we can eliminate the inertial state and explicitly formulate the relative velocity and translation as
\begin{align}
	\dot{\mathbf{t}}_{F \mid Lj}^{L}=&\Delta {\mathrm{R}_{Li j}^\top}\left(\mathrm{R}_{Fi}^{L} \Delta \dot{\mathbf{t}}_{Fi j}-\Delta \dot{\mathbf{t}}_{Li j}+\dot{\mathbf{t}}_{F \mid Li}^{L}+\boldsymbol{\omega}_{Li} ^{\wedge} \mathbf{t}_{F \mid Li}^{L}\right)-\nonumber\\
	&\boldsymbol{\omega}_{Lj}^{\wedge} \mathbf{t}_{F \mid Lj}^{L},\label{eq20}\\
	\mathbf{t}_{F \mid Lj}^{L}=&\Delta {\mathrm{R}_{Li j}^\top}\left(\mathrm{R}_{Fi}^{L} \Delta \mathbf{t}_{Fi j}-\Delta \mathbf{t}_{Li j}+\mathbf{t}_{F \mid Li}^{L}+\dot{\mathbf{t}}_{F \mid Li}^{L}T\right)+\nonumber\\
	&\Delta {\mathrm{R}_{Li j}^\top}\boldsymbol{\omega}_{Li}^\wedge \mathbf{t}_{F \mid Li}^{L}T.\label{eq21}
\end{align}
Notice that the angular velocity at instance $j$ is involved.

The IMU preintegration and angular velocity above is corrupted by noise. They are introduced to the relative state through propagation. The uncertainty propagation is derived by extracting the noisy items and performing the first-order approximation to obtain the information matrix of the factor. For the extended attitude preintegration in (\ref{eq16}), we define
\begin{equation}\label{eq22}
	\Delta \mathrm{R}_{Ci j} \doteq \Delta {\mathrm{R}_{Li j}^\top}\mathrm{R}_{Fi}^{L} \Delta \mathrm{R}_{Fi j},
\end{equation}
to analyze the noise propagation, where the capital $C$ means collaboration. We drop the frame notation and define $\mathrm{R}\doteq\mathrm{R}_F^L,\textbf{t}\doteq\textbf{t}_{F\mid L}^L,\dot{\textbf{t}}=\dot{\textbf{t}}_{F\mid L}^L$ to keep the derivation concise. Using the identical noise definition of IMU preintegration from \cite{ref10} and the Baker-Campbell-Hausdorff (BCH) formula\cite{ref12}, we obtain
\begin{align}\label{eq23}
	\Delta \mathrm{R}_{Ci j}&=\Delta \tilde{\mathrm{R}}_{Ci j} \operatorname{Exp}\left(\Delta \tilde{\mathrm{R}}_{Ci j}^\top \delta \boldsymbol{\phi}_{Li j}-\delta \boldsymbol{\phi}_{Fi j}\right) \nonumber\\
	& \doteq \Delta \tilde{\mathrm{R}}_{Ci j} \operatorname{Exp}\left(-\delta \boldsymbol{\phi}_{Ci j}\right).
\end{align}
where 
\begin{equation}
	\Delta \tilde{\mathrm{R}}_{Ci j}\doteq\Delta \tilde{\mathrm{R}}_{Li j}^\top\mathrm{R}_{Fi}^{L} \Delta \tilde{\mathrm{R}}_{Fi j},
\end{equation}
$\delta \boldsymbol{\phi}_{F(L)i j}$ is the noise of attitude preintegration, and we defined the noise of extended attitude preintegration as $\delta\boldsymbol{\phi}_{Ci j}$. The composite operator $\operatorname{Exp}(\cdot)$ is $\text {exp}((\cdot)^{\wedge})$ and $\operatorname{Ln}(\cdot)$ is the opposite.

For the extended velocity and translation preintegration in (\ref{eq17}) and (\ref{eq18}), we define
\begin{align}
	\Delta \dot{\mathbf{t}}_{Ci j} \doteq& \mathrm{R}_{i} \Delta \dot{\mathbf{t}}_{Fi j}-\Delta \dot{\mathbf{t}}_{Li j}-\Delta \mathrm{R}_{Li j}\left(\dot{\mathbf{t}}_{j}+\boldsymbol{\omega}_{Lj}^{\wedge} \mathbf{t}_{j}\right)+\boldsymbol{\omega}_{Li}^{\wedge} \mathbf{t}_{i},\label{eq24}\\
	\Delta \mathbf{t}_{Ci j} \doteq& \mathrm{R}_{i} \Delta \mathbf{t}_{Fi j}-\Delta \mathbf{t}_{Li j}-\Delta \mathrm{R}_{Li j} \mathbf{t}_{j}+\boldsymbol{\omega}_{Li}^{\wedge} \mathbf{t}_{i} T.\label{eq27}
\end{align}

Regard the noise of velocity and translation preintegration $\delta \dot{\mathbf{t}}_{F(L)i j}$ and $\delta \mathbf{t}_{F(L)i j}$, and drop the higher order items to obtain
\begin{align}
	\Delta \dot{\mathbf{t}}_{Ci j}=&\Delta \tilde{\dot{\mathbf{t}}}_{Ci j}-\mathrm{R}_{i} \delta \dot{\mathbf{t}}_{Fi j}+\delta \dot{\mathbf{t}}_{Li j} +\mathbf{t}_{i}^{\wedge} \boldsymbol{\eta}_{L g vi}-\nonumber\\
	&\Delta  \tilde{\mathrm{R}}_{Li j} \mathbf{t}_{j}^{\wedge} \boldsymbol{\eta}_{L g vj}-\Delta  \tilde{\mathrm{R}}_{Li j}\left(\dot{\mathbf{t}}_{j}+\hat{\boldsymbol{\omega}}_{Lj}^{\wedge} \mathbf{t}_{j}\right){}^{\wedge} \delta \boldsymbol{\phi}_{Li j}\nonumber\\
	\doteq & \Delta \tilde{\dot{\mathbf{t}}}_{Ci j}-\delta \dot{\mathbf{t}}_{Ci j},\label{eq26}\\
	\Delta \mathbf{t}_{Ci j}=&\Delta \tilde{\mathbf{t}}_{Ci j}-\mathrm{R}_{i} \delta \mathbf{t}_{Fi j}+\delta \mathbf{t}_{Li j}-\Delta \tilde{\mathrm{R}  }_{Li j} \mathbf{t}_{j}^{\wedge} \delta \boldsymbol{\phi}_{Li j}\nonumber\\
&+\mathbf{t}_{i}^{\wedge} \boldsymbol{\eta}_{L g vi} T\nonumber \\
\doteq&\Delta \tilde{\mathbf{t}}_{Ci j}-\delta \mathbf{t}_{Ci j},\label{eq28}
\end{align}
where
\begin{align}
	\hat{\boldsymbol{\omega}}_{Li}=&\tilde{\boldsymbol{\omega}}_{Li}-\hat{\boldsymbol{\beta}}_{L gi},\label{eq25}\\
	\Delta \tilde{\dot{\mathbf{t}}}_{Ci j} \doteq& \mathrm{R}_{i} \Delta \tilde{\dot{\mathbf{t}}}_{Fi j}-\Delta \tilde{\dot{\mathbf{t}}}_{Li j}-\Delta \tilde{\mathrm{R}}_{Li j}\left(\dot{\mathbf{t}}_{j}+\hat{\boldsymbol{\omega}}_{Lj}^{\wedge} \mathbf{t}_{j}\right)+\hat{\boldsymbol{\omega}}_{Li}^{\wedge} \mathbf{t}_{i},\label{eq26_1}\\
	\Delta \tilde{\mathbf{t}}_{Ci j} \doteq& \mathrm{R}_{i} \Delta \tilde{\mathbf{t}}_{Fi j}-\Delta \tilde{\mathbf{t}}_{Li j}-\Delta \tilde{\mathrm{R}}_{Li j} \mathbf{t}_{j}+\hat{\boldsymbol{\omega}}_{Li}^{\wedge} \mathbf{t}_{i} T,\label{eq26_2},
\end{align}
and we define $\delta \dot{\mathbf{t}}_{Ci j}$ and $\delta \mathbf{t}_{Ci j}$ as error of the extended velocity and translation preintegration.

The above uncertainty propagation expressed in a compact matrix form is
\begin{equation}\label{eq29}
	\boldsymbol{\eta}_{Ci j}^{\Delta}=\mathrm{K}_{F} \boldsymbol{\eta}_{Fi j}^{\Delta}+\mathrm{K}_{L} \boldsymbol{\eta}_{Li j}^{\Delta}+\mathrm{K}_{L g} \bar{\boldsymbol{\eta}}_{L g v}
\end{equation}
where $\boldsymbol{\eta}_{Bi j}^{\Delta} \doteq[\delta \boldsymbol{\phi}_{Bi j},\delta \dot{\mathbf{t}}_{Bi j},\delta \mathbf{t}_{Bi j}](B=F, L, C)$, $\bar{\boldsymbol{\eta}}_{L g v} \doteq[\boldsymbol{\eta}_{L g vi}, \boldsymbol{\eta}_{L g vj}]$, $\mathrm{K}_{F}, \mathrm{K}_{L} \text { and } \mathrm{K}_{L g}$ can be clearly derived from (\ref{eq23}), (\ref{eq26}) and (\ref{eq28}). The covariance of $\bar{\boldsymbol{\eta}}_{L g v}$ is $\sigma_{Lgv}^2\mathrm{I}_{6\times6}$ and we denote the covariance of the IMU preintegration of the follower and the leader as $\Sigma_{Fi j}$ and $\Sigma_{Li j}$, which can be computed iteratively\cite{ref10}. Then the covariance of $\boldsymbol{\eta}_{Ci j}^{\Delta}$ is given as
\begin{equation}\label{eq30}
	\Sigma_{Ci j}=\mathrm{K}_{F} \Sigma_{Fi j} \mathrm{K}_{F}^\top+\mathrm{K}_{L} \Sigma_{Li j} \mathrm{K}_{L}^\top+\sigma_{Lgv}^2\mathrm{K}_{L g} \mathrm{K}_{L g}^\top+\mathrm{S}+\mathrm{S}^\top,
\end{equation}
where
\begin{equation}\label{eq31}
	\mathrm{S}=\mathrm{K}_{L} E\left(\boldsymbol{\eta}_{Li j}^{\Delta} { \bar{\boldsymbol{\eta}}_{L g v}}^\top\right) \mathrm{K}_{L g}^\top,
\end{equation}
$E(\cdot)$ is the expectation operator. We drop the subscript $L$ for conciseness in the following derivation. 

For the noise $\delta \boldsymbol{\phi}_{i j}$ of attitude preintegration, $\boldsymbol{\eta}_{g vi}$ only relate to the first angular velocity, so the covariance is
\begin{equation}\label{eq32}
	E\left(\delta \boldsymbol{\phi}_{i j} \boldsymbol{\eta}_{g vi}^\top\right)=\Delta \tilde{\mathrm{R}}_{i j}^\top \Delta \tilde{\mathrm{R}}_{i(i+1)} J_{r}^{i} \sigma_{\mathrm{g} v}^{2} \Delta t,
\end{equation}
where $\Delta t$ is the duration of single inertial measurement. For the noise $\delta \dot{\mathbf{t}}_{i j}$ of velocity preintegration, $\boldsymbol{\eta}_{a v}$ and  $\boldsymbol{\eta}_{g v}$ are uncorrelated, so
\begin{equation}\label{eq33}
	E\left(\delta \dot{\mathbf{t}}_{i j} \boldsymbol{\eta}_{g vi}^{\top}\right)=\sum_{k=i}^{j-1}-\Delta \tilde{\mathrm{R}}_{i k}\left(\tilde{\mathbf{a}}_{k}-\boldsymbol{\beta}_{ai}\right)^{\wedge} E\left(\delta \boldsymbol{\phi}_{i k} \boldsymbol{\eta}_{g vi}^\top\right) \Delta t.
\end{equation}
Substitute (\ref{eq32}) in (\ref{eq33}) and notice that $E\left(\delta \boldsymbol{\phi}_{i k} \boldsymbol{\eta}_{g vi}^\top\right)=0$ for $k=i$, we obtain
\begin{equation}\label{eq34}
	E\left(\delta \dot{\mathbf{t}}_{i j} \boldsymbol{\eta}_{g vi}^\top\right)=\left(\Delta \tilde{\dot{\mathbf{t}}}_{i(i+1)}-\Delta \tilde{\dot{\mathbf{t}}}_{i j}\right)^{\wedge} \Delta \tilde{\mathrm{R}}_{i(i+1)} J_{r}^{i} \sigma_{\mathrm{g} v}^{2} \Delta t
\end{equation}

For the noise $\delta \mathbf{t}_{i j}$ of translation preintegration, the covariance with $\boldsymbol{\eta}_{g vi}$ is
\begin{align}\label{eq35}
	E\left(\delta \mathbf{t}_{i j} \boldsymbol{\eta}_{g vi}^\top\right)&=\sum_{k=i}^{j-1}[E\left(\delta \dot{\mathbf{t}}_{i k} \boldsymbol{\eta}_{g vi}^\top\right) \Delta t-\\
	&\frac{1}{2} \Delta \tilde{\mathrm{R}}_{i k}\left(\tilde{\mathbf{a}}_{k}-\boldsymbol{\beta}_{ai}\right)^{\wedge} E\left(\delta \boldsymbol{\phi}_{i k} \boldsymbol{\eta}_{g vi}^\top\right) \Delta t^{2}]\nonumber
\end{align}
Substitute (\ref{eq32}) and (\ref{eq34}) into (\ref{eq35}), we obtain
\begin{align}\label{eq36}
	E&\left(\delta \mathbf{t}_{i j} \boldsymbol{\eta}_{g vi}^\top\right)=\left(\Delta \tilde{\mathbf{t}}_{i(i+1)}-\Delta \tilde{\mathbf{t}}_{i j}\right){}^{\wedge} \Delta \tilde{\mathrm{R}}_{i(i+1)} J_{r}^{i} \sigma_{\mathrm{g} v}^{2} \Delta t \nonumber\\
	&+(j-i-1) \Delta \tilde{\dot{\mathbf{t}}}_{i(i+1)}^{\wedge} \Delta \tilde{\mathrm{R}}_{i(i+1)} J_{r}^{i} \sigma_{\mathrm{g} v}^{2} \Delta t.
\end{align}
Finally substitute (\ref{eq32}), (\ref{eq34}) and (\ref{eq36}) into (\ref{eq31}) to calculate $\mathrm{S}$ and further the covariance of the extended preintegration. Because $\boldsymbol{\eta}_{Bi j}^{\Delta}$ is the linear combination of zero-mean and Gaussian error $\boldsymbol{\eta}_{Fi j}^{\Delta}$, $\boldsymbol{\eta}_{Li j}^{\Delta}$, and $\bar{\boldsymbol{\eta}}_{L g v}$, it is zero-mean and Gaussian up to first-order and the information matrix is the inversion of (\ref{eq30}).
%
%
%
%
%
%
%

\subsection{Residual and Jacobian}
We define residual of the extended attitude, velocity, and translation preintegration based on (\ref{eq16}), (\ref{eq17}), (\ref{eq18}) as
\begin{align}\label{eq4_25}
	\mathbf{r}_{\boldsymbol{\phi}_{Ci j}}\doteq&\operatorname{Ln}(\mathrm{R}_{j}^\top \Delta \bar{\mathrm{R}}_{Li j}^\top\mathrm{R}_{i}\Delta \bar{\mathrm{R}}_{Fi j}),\\
	\mathbf{r}_{\dot{\mathbf{t}}_{Ci j}}\doteq& \mathrm{R}_{i}\Delta \bar{\dot{\mathbf{t}}}_{Fi j}-\Delta \bar{\dot{\mathbf{t}}}_{Li j}-\Delta \bar{\mathrm{R}}_{Li j}\left(\dot{\mathbf{t}}_{j}+\bar{\boldsymbol{\omega}}_{Lj}^{\wedge} \mathbf{t}_{j}\right)+\nonumber\\
	&\left(\dot{\mathbf{t}}_{i}+\bar{\boldsymbol{\omega}}_{Li}^{\wedge} \mathbf{t}_{i}\right),\nonumber\\
	\mathbf{r}_{\mathbf{t}_{Ci j}}\doteq&\mathrm{R}_{i}\Delta \bar{\mathbf{t}}_{Fi j}-\Delta \bar{\mathbf{t}}_{Li j}-\Delta \bar{\mathrm{R}}_{Li j}\mathbf{t}_{j}+\mathbf{t}_{i}+\left(\dot{\mathbf{t}}_{i}+\bar{\boldsymbol{\omega}}_{Li}^{\wedge} \mathbf{t}_{i}\right) T,\nonumber
\end{align}
where $\Delta \bar{\mathrm{R}}_{Bi j}$, $\Delta \bar{\dot{\mathbf{t}}}_{Bi j}$ and $\Delta \bar{\mathbf{t}}_{Bi j}(B=F,L)$ are the first-order update of IMU preintegration with biases perturbation proposed in \cite{ref10} and 
\begin{equation}
	\bar{\boldsymbol{\omega}}_{L}\doteq\tilde{\boldsymbol{\omega}}_{L}-\hat{\boldsymbol{\beta}}_{Bg}-\delta \boldsymbol{\beta}_{Bg}\doteq\hat{\boldsymbol{\omega}}_{L}-\delta \boldsymbol{\beta}_{Bg}.
\end{equation}
The states of interest are the relative pose, velocity, and biases of both IMUs at instances $i$ and $j$. We use $\mathrm{R}=\widehat{\mathrm{R}} \operatorname{Exp}(\delta \alpha)$ for attitude update and $\mathbf{y}=\widehat{\mathbf{y}}+\delta\mathbf{y}$ for the rest. 

The residual of extended attitude preintegration only relates to the relative attitude at both instances and the gyroscope biases of both IMUs at instant $i$, so the relevant Jacobian entries are
\begin{align*}
	&\frac{\partial \mathbf{r}_{\boldsymbol{\phi}_{Ci j}}}{\partial \delta \boldsymbol{\alpha}_{i}}=\mathrm{J}_{r}^{-1}\left(\mathbf{r}_{\boldsymbol{\phi}_{Ci j}}\right) \Delta \tilde{\mathrm{R}}_{Fi j}^\top, \frac{\partial \mathbf{r}_{\delta \boldsymbol{\varphi}_{i j}}}{\partial \delta \boldsymbol{\alpha}_{j}}=-\mathrm{J}_{r}^{-1}\left(-\mathbf{r}_{\delta \boldsymbol{\varphi}_{i j}}\right), \\
	&\frac{\partial \mathbf{r}_{\boldsymbol{\phi}_{Ci j}}}{\partial \delta {\boldsymbol{\beta}}_{Fgi}}=\mathrm{J}_{r}^{-1}\left(\mathbf{r}_{\boldsymbol{\phi}_{Ci j}}\right) \frac{\partial \Delta {\mathrm{R}}_{Fi j}}{\partial \boldsymbol{\beta}_{Fgi}},\\
	&\frac{\partial \mathbf{r}_{\boldsymbol{\phi}_{Ci j}}}{\partial \delta {\boldsymbol{\beta}}_{Lgi}}=-\mathrm{J}_{r}^{-1}\left(\mathbf{r}_{\boldsymbol{\phi}_{Ci j}}\right)\hat{\mathrm{R}}_{j}^\top\frac{\partial \Delta {\mathrm{R}}_{Li j}}{\partial \boldsymbol{\beta}_{Lgi}}
\end{align*}
The partial derivatives of $\mathbf{r}_{\boldsymbol{\phi}_{Ci j}}$ w.r.t. other states are zeros.

The extended velocity preintegration is unrelated to the attitude at instance $j$ and the rest of associated Jacobian entries are
\begin{align*}
	&\frac{\partial \mathbf{r}_{\dot{\mathbf{t}}_{i j}}}{\partial \delta \boldsymbol{\alpha}_{i}}=-\mathrm{R}_{i} \Delta \tilde{\dot{\mathbf{t}}}_{Fi j}^{\wedge}, \frac{\partial \mathbf{r}_{\dot{\mathbf{t}}_{i j}}}{\partial \delta \mathbf{t}_{i}}=\hat{\boldsymbol{\omega}}_{Li}^{\wedge}, \frac{\partial \mathbf{r}_{\dot{\mathbf{t}}_{i j}}}{\partial \delta \dot{\mathbf{t}}_{i}}=\mathrm{I}_{3 \times 3},\\
	&\frac{\partial \mathbf{r}_{\dot{\mathbf{t}}_{i j}}}{\partial \delta \mathbf{t}_{j}}=-\Delta \tilde{\mathrm{R}}_{Li j}\hat{\boldsymbol{\omega}}_{Lj}^{\wedge},
	\frac{\partial \mathbf{r}_{\dot{\mathbf{t}}_{i j}}}{\partial \delta \dot{\mathbf{t}}_{j}}=-\Delta \tilde{\mathrm{R}}_{Li j}, \\
	&\frac{\partial \mathbf{r}_{\dot{\mathbf{t}}_{i j}}}{\partial \delta {\boldsymbol{\beta}}_{Fgi}}=\mathrm{R}_{i} \frac{\partial \Delta \dot{\mathbf{t}}_{Fi j}}{\partial \boldsymbol{\beta}_{Fgi}}, \frac{\partial \mathbf{r}_{\dot{\mathbf{t}}_{i j}}}{\partial \delta {\boldsymbol{\beta}}_{Fai}}=\mathrm{R}_{i} \frac{\partial \Delta \dot{\mathbf{t}}_{Fi j}}{\partial \boldsymbol{\beta}_{Fai}}\\
	&\frac{\partial \mathbf{r}_{\dot{\mathbf{t}}_{i j}}}{\partial \delta {\boldsymbol{\beta}}_{Lgi}}=-\frac{\partial \Delta \dot{\mathbf{t}}_{Li j}}{\partial \boldsymbol{\beta}_{Lgi}}+\Delta \tilde{\mathrm{R}}_{Li j}\left(\dot{\mathbf{t}}_{j}+
	\hat{\boldsymbol{\omega}}_{Lj}^{\wedge} \mathbf{t}_{j}\right)^{\wedge}\frac{\partial \Delta {\mathrm{R}}_{Li j}}{\partial \boldsymbol{\beta}_{Lgi}} + \mathbf{t}_{i}^{\wedge}\\
	&\frac{\partial \mathbf{r}_{\dot{\mathbf{t}}_{i j}}}{\partial \delta {\boldsymbol{\beta}}_{Lai}}=-\frac{\partial \Delta \dot{\mathbf{t}}_{Li j}}{\partial \boldsymbol{\beta}_{Lai}}, \frac{\partial \mathbf{r}_{\dot{\mathbf{t}}_{i j}}}{\partial \delta {\boldsymbol{\beta}}_{Lgj}}=-\Delta \tilde{\mathrm{R}}_{Li j}\mathbf{t}_{j}^{\wedge}
\end{align*}

The residual of extended translation preintegration is irrelevant to the relative attitude and velocity at instant $j$, so associate entries in the Jacobian matrix are zero and the others entries are
\begin{align*}
	&\frac{\partial \mathbf{r}_{\mathbf{t}_{i j}}}{\partial \delta \boldsymbol{\alpha}_{i}}=-\mathrm{R}_{i} \Delta \tilde{\mathbf{t}}_{Fi j}^{\wedge}, \frac{\partial \mathbf{r}_{\mathbf{t}_{i j}}}{\partial \delta \mathbf{t}_{i}}=\mathrm{I}_{3 \times 3}+\tilde{\boldsymbol{w}}_{Li}^{\wedge} \Delta t_{i j}, \\ &\frac{\partial \mathbf{r}_{\mathbf{t}_{i j}}}{\partial \delta \dot{\mathbf{t}}_{i}}=\mathrm{I}_{3 \times 3} \Delta t_{i j},
	\frac{\partial \mathbf{r}_{\mathbf{t}_{i j}}}{\partial \delta \mathbf{t}_{j}}=-\Delta \tilde{\mathrm{R}}_{Li j}, \\
	&\frac{\partial \mathbf{r}_{\mathbf{t}_{i j}}}{\partial \delta {\boldsymbol{\beta}}_{Fgi}}=\mathrm{R}_{i} \frac{\partial \Delta \mathbf{t}_{Fi j}}{\partial \boldsymbol{\beta}_{Fgi}}, \frac{\partial \mathbf{r}_{\mathbf{t}_{i j}}}{\partial \delta {\boldsymbol{\beta}}_{Fai}}=\mathrm{R}_{i} \frac{\partial \Delta \mathbf{t}_{Fi j}}{\partial \boldsymbol{\beta}_{Fai}},\\
	&\frac{\partial \mathbf{r}_{\mathbf{t}_{i j}}}{\partial \delta {\boldsymbol{\beta}}_{Lgi}}=-\frac{\partial \Delta \mathbf{t}_{Li j}}{\partial \boldsymbol{\beta}_{Lgi}}+\Delta \tilde{\mathrm{R}}_{Li j}\mathbf{t}_{j}^{\wedge}\frac{\partial \Delta {\mathrm{R}}_{Li j}}{\partial \boldsymbol{\beta}_{Lgi}}+\mathbf{t}_{i}^{\wedge}\Delta t_{i j}\\
	&\frac{\partial \mathbf{r}_{\mathbf{t}_{i j}}}{\partial \delta {\boldsymbol{\beta}}_{Lai}}=-\frac{\partial \Delta \mathbf{t}_{Li j}}{\partial \boldsymbol{\beta}_{Lai}}
\end{align*}

With the residual, the analytical Jacobian, and the information matrix, the extended preintegration can serve as a factor in the relative state estimator. 

\subsection{Algorithm Design}
The relative state estimator shown in Fig.\ref{fig_2} and summarized in Algorithm \ref{alg1} is based on nonlinear optimization.
The camera on the leader generates the image with the perspective of the marks fixed on the follower. Visual features $\varTheta_j$ are extracted from these images for initialization using PNP solver\cite{ref14} or serve as features factor in the nonlinear optimization. The inertial measurements of both platforms between two successive image are pre-integrated to obtain the IMU preintegration $\mathcal{I}_{F(L)ij}$. 
Both IMU preintegration are combined as the extended preintegration $\mathcal{I}_{Cij}$ and serve as another factor in the optimization. The prior state, random walk constraint of both IMUs, and the above two factors are fused in the estimator to calculate the optimal relative state in the tracking thread. Alternatively, the refinement thread can be launched for a full smoother to incorporate all feature factors, extended preintegration, and random walk constraint on the entire trajectory as the typical process in SLAM\cite{ref15}.
\begin{algorithm}[H]
	\caption{Relative state estimator}\label{alg:alg1}
	\renewcommand{\algorithmicrequire}{\textbf{Input:}}
	\renewcommand{\algorithmicensure}{\textbf{Output:}}
	\begin{algorithmic}[1]
		\REQUIRE Image at instant $j$, IMU measurement between instance $i$ and $j$ of both platforms, and the prior state $X_i$  
		\ENSURE Relative state $X_j$    
		\STATE Extract marks $\varTheta_j$ from the image
		\IF {Uninitialized}
		\STATE Run PNP solver with $\varTheta_j$ for initialization 
		\ELSE
		\STATE Calculate the IMU preintegration $\mathcal{I}_{Fij}$ and $\mathcal{I}_{Lij}$
		\STATE Calculate the extended preintegration $\mathcal{I}_{ij}$ 
		\STATE Nonlinear optimization with $\mathcal{I}_{ij}$ and $\varTheta_j$
		\STATE Marginalize the relative state $X_i$
		\ENDIF
	\end{algorithmic}
	\label{alg1}
\end{algorithm}
%
%

\section{Experiment}\label{sec3}
This section performs two experiments to verify the proposed algorithms on simulated and real data. Section\ref{sec4_1} compares the proposed algorithms against two filter-based methods from the aspects of trajectory dynamics and image recognition rates. Section\ref{sec4_2} uses the proposed estimator to realize the visual tracking of the 6DoF controller in VR application.



\subsection{Simulation}\label{sec4_1}
\begin{figure}[!t]
	\centering
	\includegraphics[width=3.0in]{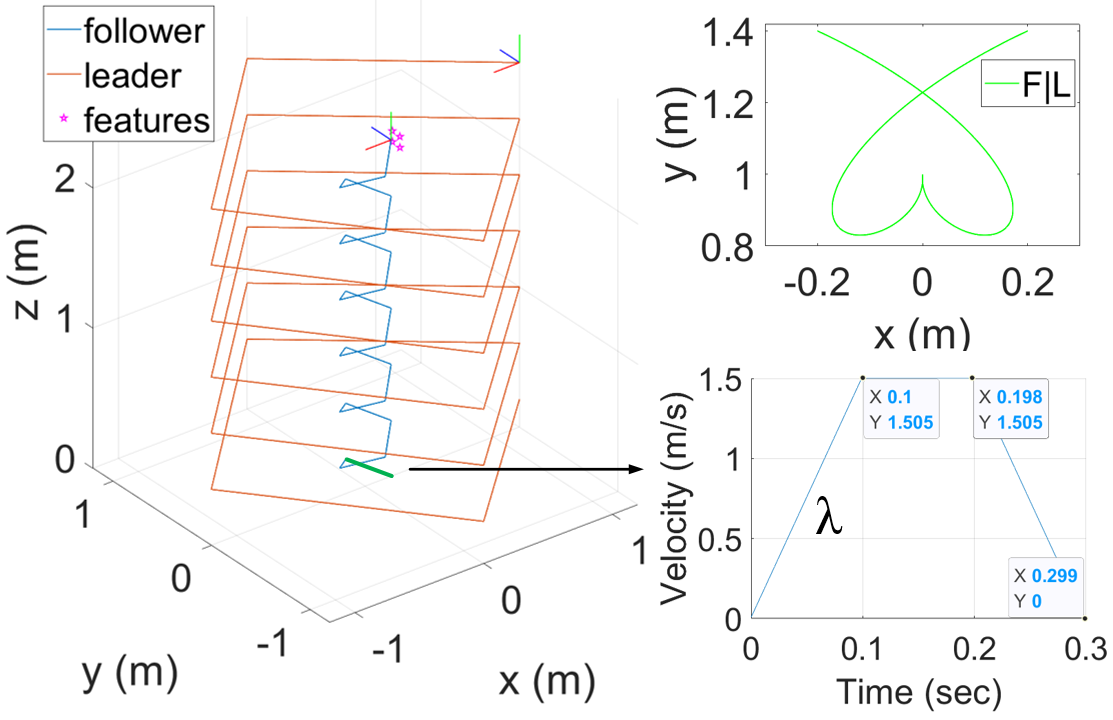}
	\caption{Left: the inertial trajectory of both platforms with detected visual features. Upper right: relative trajectory at the xy plane. Bottom right: the velocity of the first linear trajectory of the follower with acceleration amplitude $\lambda$.}
	\label{fig_3}
\end{figure}
Two filter-based algorithms with different relative kinematics are included for comparison. The first kinematics calculates numerical integration on the relative frame with the relative acceleration, which is composed of the Euler acceleration calculated with applied torque or angular acceleration \cite{ref0}; the second kinematics performs integration on the inertial frame and converts the result to the relative frame to avoid involving the Euler acceleration\cite{ref9}. Because the applied torque is inaccessible as we introduce in section \ref{sec1}, the Euler acceleration in the first kinematics is approximatively calculated with differential angular velocity measurements. According to the highest order of relative states involved in the kinematics, the algorithms are denoted as 3rdOrderEKF and 2ndOrderEKF in the following. The original algorithms rely on a perfect IMU measurement from the leader with known biases and noise eliminated, which is impractical. For a fair comparison, the noise propagation is derived and included in both algorithms, and the biases are fixed as zeros. Accordingly, we fix the biases of the leader’s IMU as zeros in our relative state estimator for another comparison. The estimators with and without the above biases are denoted as FullOpt and MinorOpt. Both estimators only perform one iteration with the Gauss-Newton solver. Besides, we add the full smoother to the MinorOpt and denote the algorithm as GBAOpt.

The comparisons are performed from two perspectives, i.e., the trajectory dynamic and image recognition rate $\gamma$ defined as the percentage of images with detected visual features. We run 100 Monte-Carlo simulations with normally distributed initial state error. The standard deviation of the initial relative attitude, translation, velocity, and biases of the gyro and accelerometer are 0.02rad, 0.02m, 0.2m/s, 0.01rad/s, and $0.05\mathrm{m/s^2}$. The performance is evaluated by the root-mean-square error RMSE$\left(\mathbf{q}\right)$, where the error vector $\mathbf{q}$ including the relative attitude, translation, and velocity are defined as $\delta \boldsymbol{\theta}=\textrm{Ln}(\hat{\textrm{R}}_{F}^{L\top} \textrm{R}_{F}^{L})$, $\delta \mathbf{p}=\mathbf{t}_{F\mid L}^L-\mathbf{\hat{t}}_{F\mid L}^L$ and  $\delta \mathbf{v}=\dot{\mathbf{t}}_{F\mid L}^L-\hat{\dot{\mathbf{t}}}_{F\mid L}^L$.

\begin{figure*}[!t]
	\centering
	\includegraphics[width=6.5in]{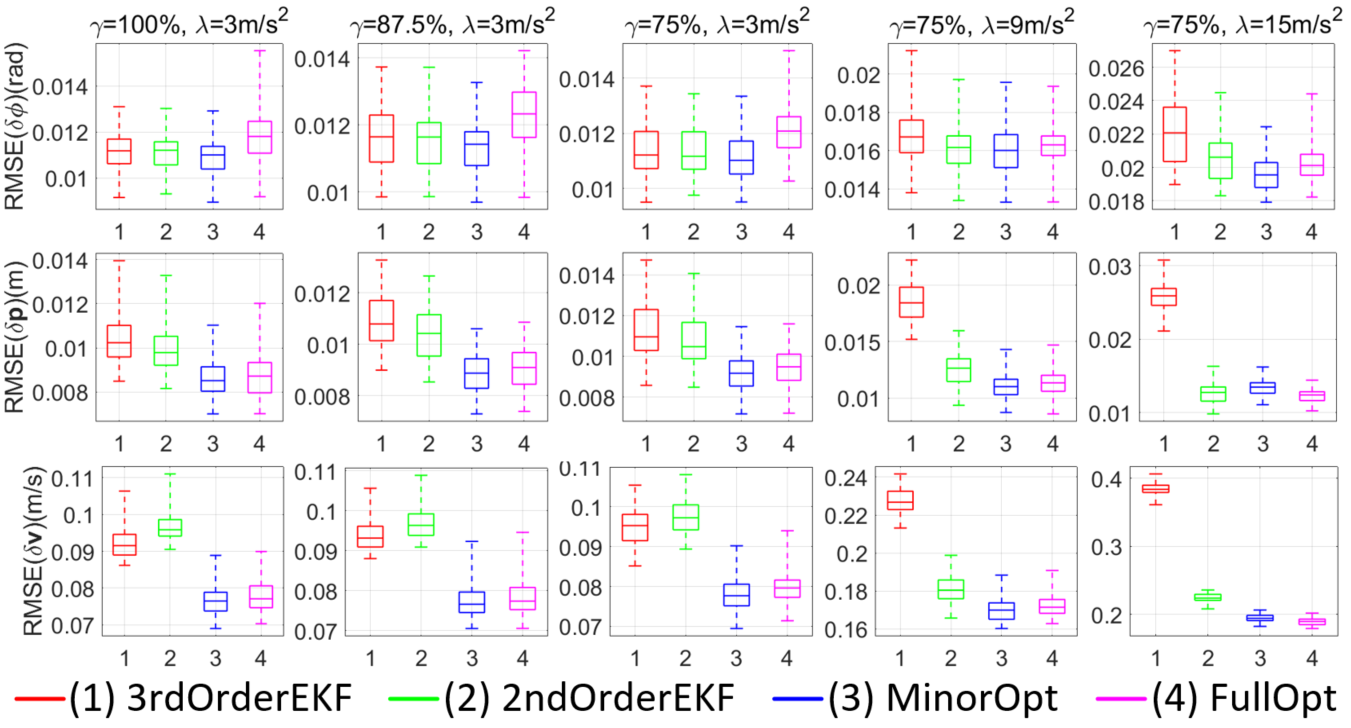}
	\caption{Algorithms comparisons with RMSE of the relative attitude, position and velocity .}
	\label{fig_5}
\end{figure*}
\begin{table}[!t]
	\caption{IMU Continuous-Time Noise Parameters of the leader and follower platforms\label{tab1}}
	\centering
	\begin{tabular}{|c||c||c|}
		\hline
		& leader & follower\\
		\hline
		$\sigma_{g v}\left[\mathrm{rad} /(s \sqrt{\mathrm{Hz}})\right]$ & $1.528 \times 10^{-3}$ & $2.269 \times 10^{-3}$\\
		\hline
		$\sigma_{g u}\left[\mathrm{rad} /\left(s^{2} \sqrt{\mathrm{Hz}}\right)\right]$ & $1.867 \times 10^{-5}$ & $1.536 \times 10^{-5}$\\
		\hline
		$\sigma_{a v}\left[\mathrm{m} /\left(s^{2} \sqrt{\mathrm{Hz}}\right)\right]$ & $1.244 \times 10^{-2}$ & $8.182 \times 10^{-3}$\\
		\hline
		$\sigma_{a u}\left[\mathrm{m} /\left(s^{3} \sqrt{\mathrm{Hz}}\right)\right]$ & $7.841 \times 10^{-4}$ & $6.154 \times 10^{-4}$\\
		\hline
	\end{tabular}
\end{table}

We design the trajectory of both platforms with cyclically ascending line segments shown on the left of Fig.\ref{fig_3}. The trajectories of the follower and leader on the XY plane are squares with lengths of 0.283m and 1.414m, and both platforms elevate 0.1m during every segment. The follower rotates 45 degrees along the axis $\mathbf{p}_1=[1~0~0]^\top$, $\mathbf{p}_2=[0~1~0]^\top$ and $\mathbf{p}_3=[1~1~1]^\top$ during the first, second and third four segments, and repeat the rotation from the fourth four segments. The leader rotates $-\left[\pi/2~\pi/2~0\right]$ in the ZYX sequence of Euler angles during every segment. Each platform simultaneously arrives at every vertex. This complicated trajectory ensures a wide range of relative poses and a high image recognition rate. The velocity of the follower during the first segment is shown on the bottom right of Fig.\ref{fig_3} including uniform acceleration and deceleration successively with the same acceleration amplitude, denoted as $\lambda$. The relative trajectory is shown in the upper right of Fig.\ref{fig_3}, where the constant relative height is 20cm. We use different $\lambda$ to verify the performance in the following.

A cube with marks as the Apriltags\cite{ref13} attached on each face rigidly joints the follower. Each mark has four features on the corner of a square with 0.14m in length. A camera with a pinhole perspective projection is equipped on the leader to detect the marks when the angle between the face's normal vector and the camera's line of sight is between 120 to 240 degrees, and the four features are within the image with a size of 640 $\times $480 pixels. The images are sampled at 25Hz, and the standard deviation of features measurement is $\sigma_{\rho k}=1$ pixels. Although each image has one or two marks detected in the trajectory, we drop some images to simulate the recognition rate, i.e. $\gamma=75\%$ means we drop one in five images. This can happen in high-dynamic trajectories where some images are blurred. Besides, two IMUs with the noise parameters shown in Table \ref{tab1} are equipped on the platforms to measure linear acceleration and angular velocity at 250Hz.
\begin{figure}[!t]
	\centering
	\includegraphics[width=3.4in]{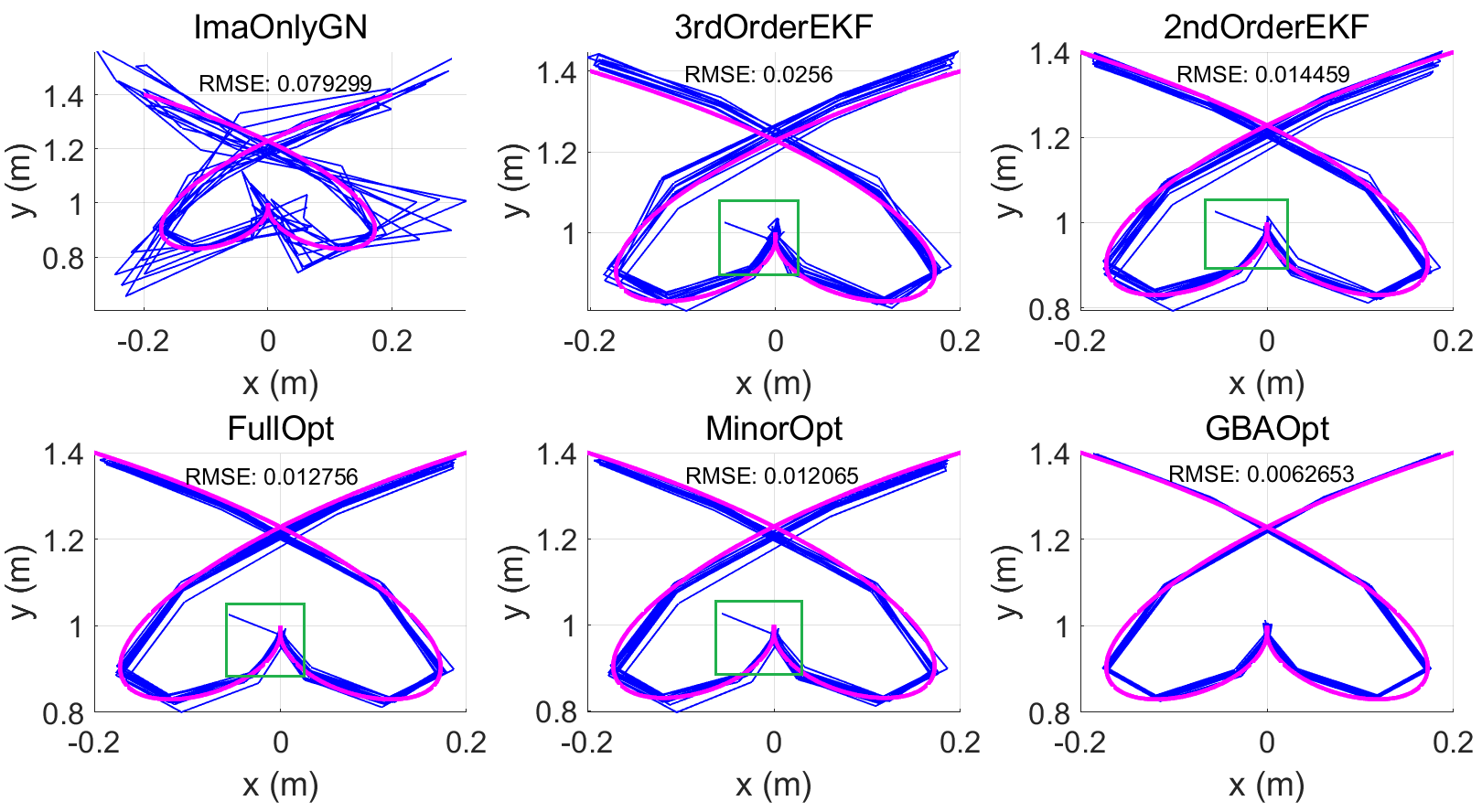}
	\caption{Estimated trajectories with ground truth. The green rectangle shows the area with large error.}
	\label{fig_6}
\end{figure}

\begin{figure}[!t]
	\centering
	\includegraphics[width=3.4in]{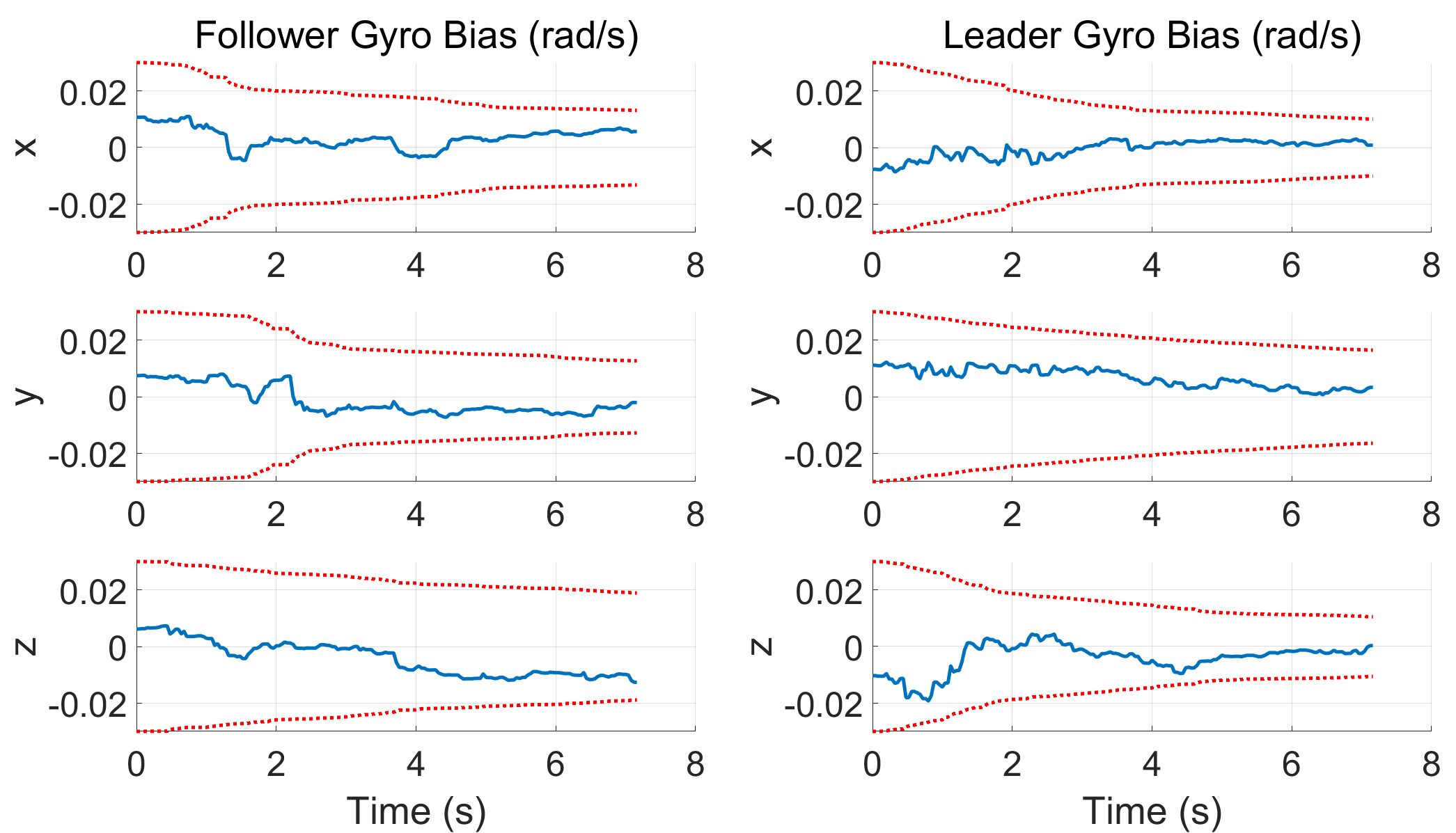}
	\caption{Estimations of the gyro biases of the Full-Opt are converged within the uncertainty boundary (3$\sigma$) with relatively large errors.}
	\label{fig_15}
\end{figure}

The performance of algorithms under different $\lambda$ and $\gamma$ is shown in Fig.\ref{fig_5}. Our algorithms achieve the best precision even in challenging situations with large $\lambda$ and small $\gamma$, especially for the relative translation and velocity. Because our algorithms jointly estimate two successive states as shown in Algorithm \ref{alg1}, they outperform in relative velocity estimation the EKF-based algorithms with only the current state involved in every estimated cycle. This advantage is more obvious in highly dynamic trajectories with large $\lambda$ where the approximated relative acceleration noticeably corrupts the 3rdOrderEKF as shown in the estimation of relative translation and velocity. Nevertheless, the 2ndOrderEKF is less sensitive to the trajectory dynamic because of the accurate relative kinematics. We further add the GBAOpt and the Gauss-Newton solver with only visual observation, denoted as ImgOnlyGN, to compare with the above four algorithms in the challenging configuration ($\lambda=15$, $\gamma=75\%$), and the estimated trajectory is shown in Fig.\ref{fig_6}. The small mark size and image recognition rate corrupt the ImgOnlyGN noticeably, even in this close-distance scenario. The result of the 3rdOrderEKF, 2ndOrderEKF, FullOpt, and MinorOpt are consistent with the above comparison. We mark the significant errors caused by linearization errors with green rectangles. Although the velocity changes sharply in this position, the GBAOpt can still generate a smooth trajectory with the specific full smoother. The precisions decrease similarly for all algorithms as the $\gamma$ decrease and the recognition failure has limited influence on ours because of the accurate relative velocity estimation. Although our algorithms give better precision of relative translation and velocity, the performances of relative attitude are similar, and jointly estimating the leader’s biases even decreases the accuracy in low-dynamic trajectory. The similarity is because the attitude propagations used in all algorithms are actually identical with numerical integration on the rotation group. Although the biases estimation of the FullOpt is converged within the uncertainty boundary (3$\sigma$) as shown in Fig. \ref{fig_15}, we acknowledge that the error is large w.r.t. the random walk variance, which causes the relatively large error of attitude.

\begin{figure}[!t]
	\centering
	\includegraphics[width=3.4in]{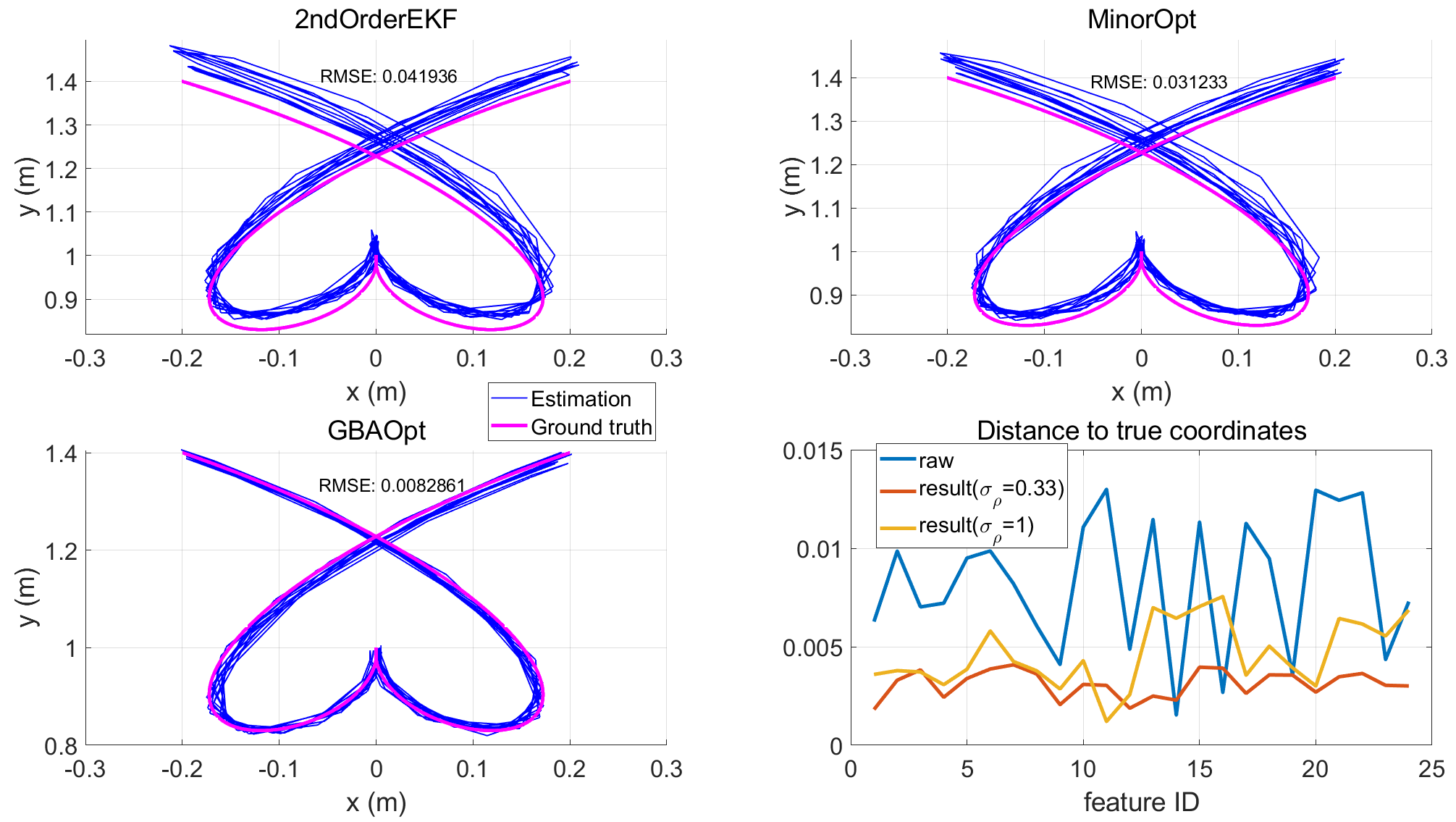}
	\caption{The ground truths and estimated trajectories of three algorithms using wrong features’ 3D coordinates. Bottom right: the distance between the true and noisy coordinate, including the raw and estimated coordinate calculated using measurement with different noises.}
	\label{fig20}
\end{figure}

Our MinorOpt performs almost the best among the four algorithms shown in Fig. \ref{fig_5} and our GBAOpt even improves the precision by 1.926 times with true features’ 3D coordinate $ \mathbf{t}_{k \mid F}^{F} $ in (\ref{ eq5_ }). However, the true coordinate is unknown in practice due to assembly inaccuracy. We further add the coordinates as variable nodes to be estimated in the full smoother with the extended preintegration, which is similar to using the IMU preintegration to optimize the 3D landmark in a global bundle adjustment \cite{ref18}, and the result is shown in Fig. \ref{fig20}. The algorithms are run with wrong coordinates generated by adding Gaussian noises with a standard deviation of 0.005m to the truth where $\lambda=9$ and $\gamma=100\%$. Among the algorithm, only the GBAOpt optimize the coordinate to obtain a precise result with an absolute trajectory error (ATE) of 8.27mm, and the rest two are corrupted heavily with cm-level ATE. With precise features measurement, we can obtain accurate estimations of the coordinates as shown in the bottom right of the figure.

\subsection{Real Experiment}\label{sec4_2}
\begin{figure}[!t]
	\centering
	\includegraphics[width=3.0in]{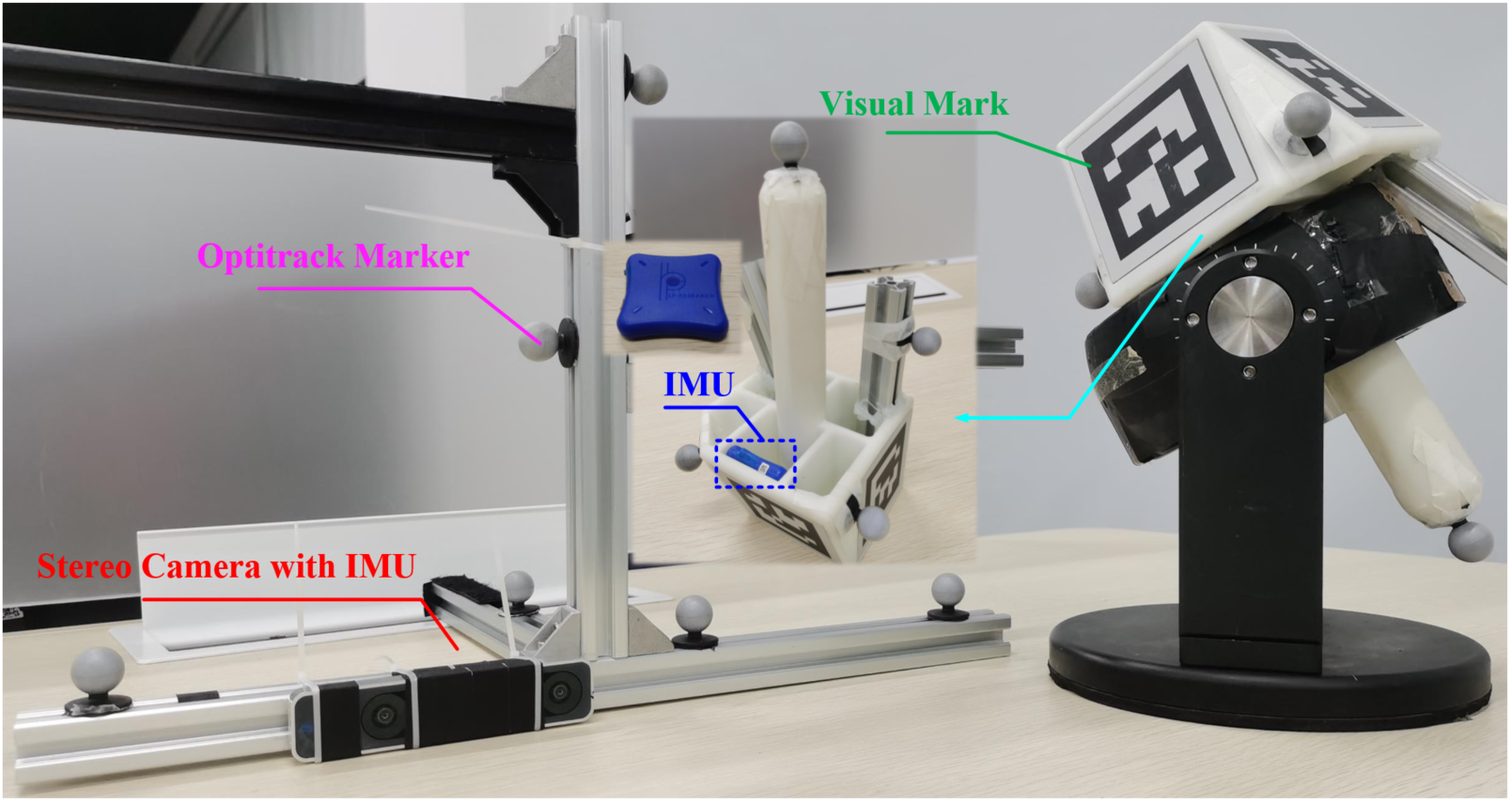}
	\caption{The 6DoF controller and headset used in our experiment.}
	\label{fig_10}
\end{figure}

This experiment uses the proposed relative state estimator to realize the visual tracking of the 6DoF controller in the VR scenario. The application requires the relative state of the controller w.r.t. the headset with multi-Camera. The controller is equipped with an IMU and visual marks to be detected. The state-of-the-art solutions use the visual features, the global state of the headset, and the inertial measurement to estimate the global state of the controller and further its relative state w.r.t. the headset. This method is susceptible to the surroundings with missing features, light mutation, dynamic scene, etc. We instead estimate the relative state directly with the proposed estimator. We use Apriltags \cite{ref13} with stable detection as marks to focus more on the estimator. The T265 visual inertia module\cite{ref16} with large FOV is Utilized. The controller is equipped with an LPMS-B2 IMU\cite{ref17}. The IMU parameter of the controller and the headset are listed in Table \ref{tab1}, and the sampling period is 5ms. The stereo camera work at 30Hz, and the FOV is 163$\deg$. To evaluate the performance, we use Optitrack to provide the ground truth. The platforms are shown in Fig.\ref{fig_10}.

\begin{figure}[!t]
	\centering
	\includegraphics[width=3.5in]{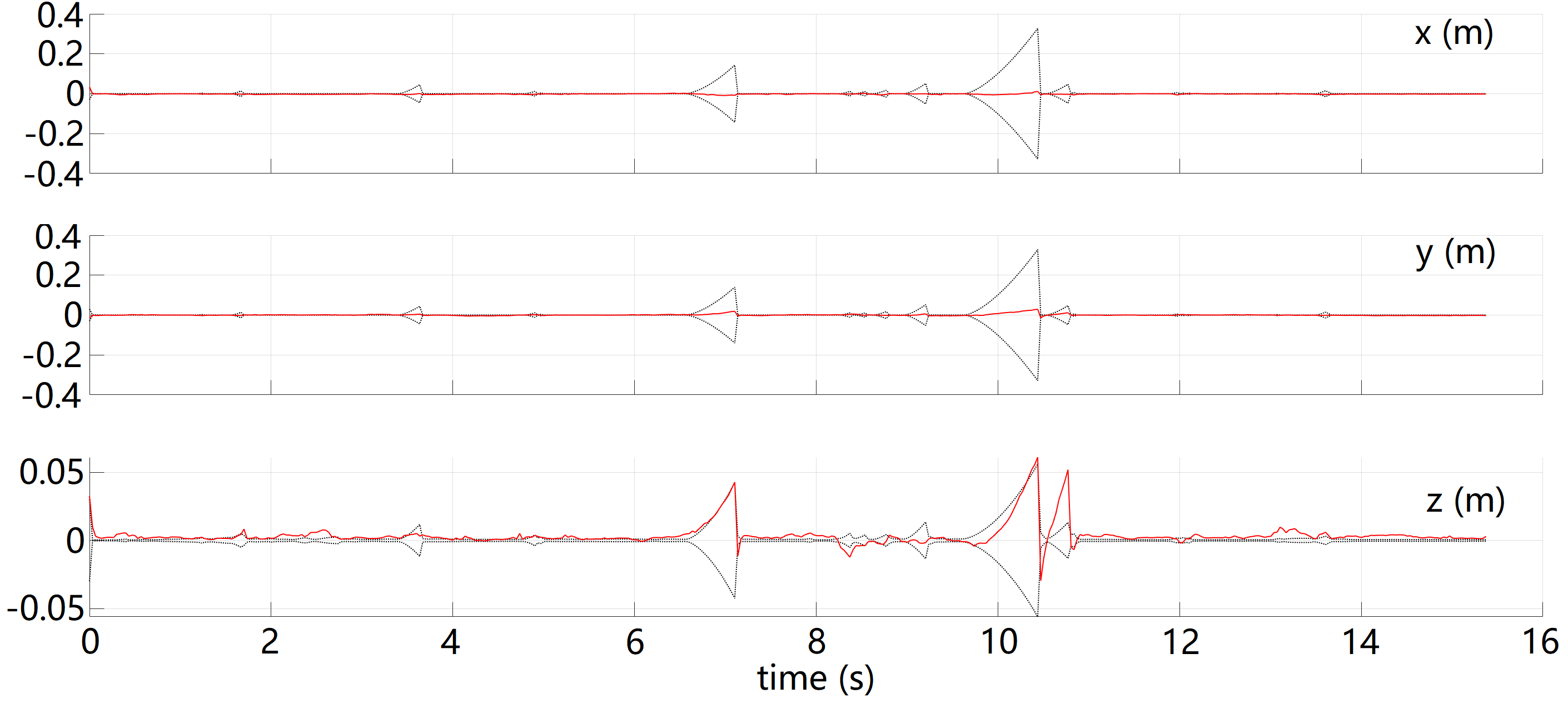}
	\caption{Localization error of the 6DoF controller w.r.t. the headset and the $3\sigma$ boundary.}
	\label{fig5_8}
\end{figure}

\begin{figure}[!t]
	\centering
	\includegraphics[width=3.5in]{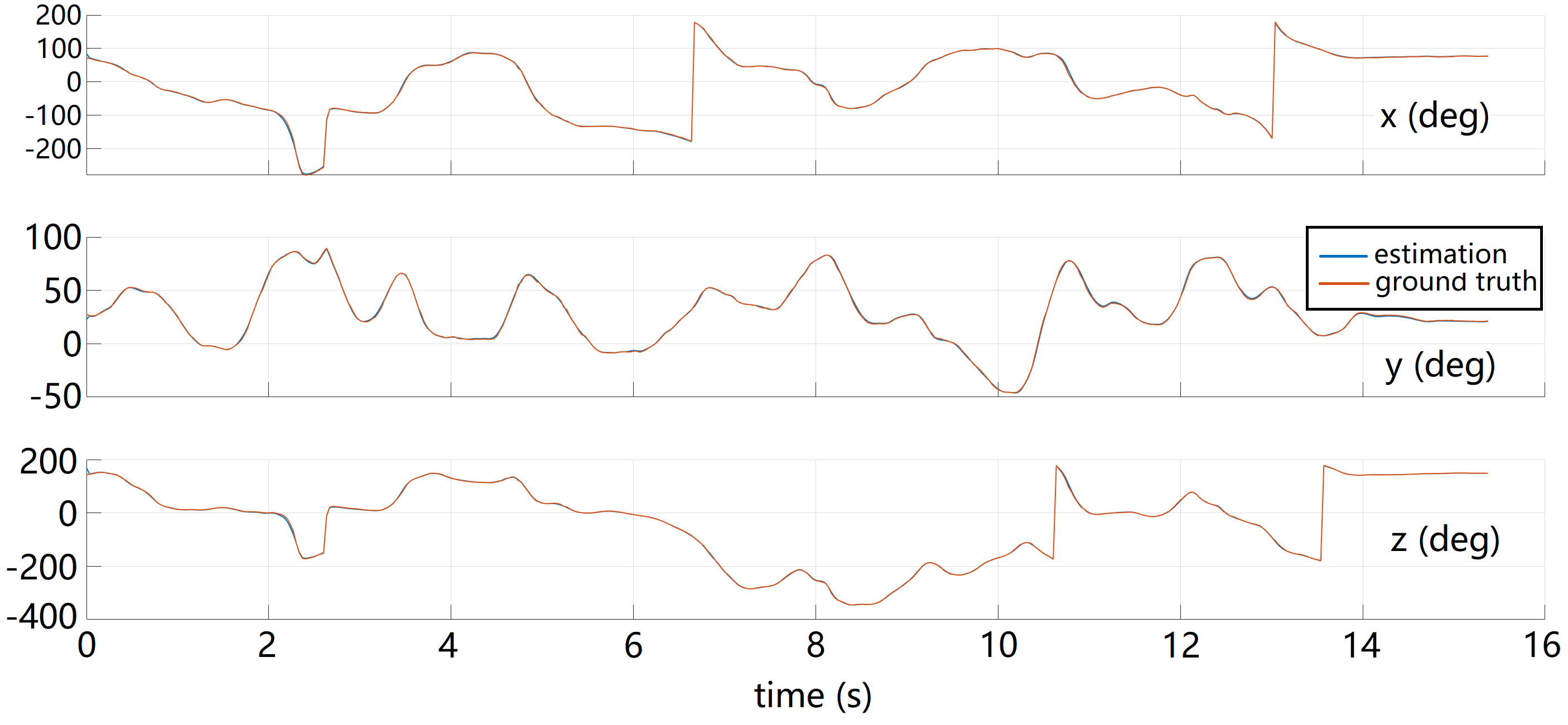}
	\caption{Attitude of the 6DoF controller w.r.t. the headset and ground truth.}
	\label{fig5_9}
\end{figure}

The controller starts at 0.7m in front of the headset and follows a circular trajectory with a 20cm radius around each axis with an angular velocity of about 117deg/s. Meanwhile, the headset rotates in a range of 30 degrees with an angular velocity of about 15deg/s. The result of relative translation is shown in Fig.\ref{fig5_8}. With detected features, the precision of relative translation is in the millimeter scale. During 6.5s to 7.2s and 9.6s to 10.9s, the detection is lost since image blur happens for the high-dynamic motion of the controller. The precision of the relative translation drops to cm-level, and the uncertainty is noticeably enlarged. Once the features are detected, the error sharply decreases to mm-level. The overall RMSE of the trajectory is 6.153 millimeters. Fig.\ref{fig5_9} shows the relative attitude estimation with ground truth. The precision is less sensitive to the detection loss because of the accurate angular velocity measurement and the overall RMSE is 1.008$\deg$. In general, except for the long duration of detection loss, the proposed 6DoF controller can provide satisfactory performance and broad applicability to almost all environments.

\section{Conclusion}\label{sec4}
This paper proposes the novel extended preintegration to combine the IMU preintegration from two collaborative platforms and design the relative state estimators (FullOpt) for real-time tracking and full smoother for refinement. We define MinorOpt as the estimator without the biases of the leader, and GBAOpt as MinorOpt with the full smoother. The MinorOpt outperforms existing methods in the simulation, and the precision is further improved significantly with the GBAOpt. The FullOpt fails to perform well in the vision-based estimator because the FOV limitation restricts the range of motion and further impacts the estimated biases. It can perform better in the ranging-based estimator, which will be verified in our future work. In the situation with wrong features 3D coordinate, most of the algorithms have large ATE except for the GBAOpt with coordinate estimation. The real experiment uses the MinorOpt to realize the 6DoF controller in the VR scenario with relative localization in the millimeter scales and well flexibility to environments, which proves the novelty of the extended preintegration.


\newpage

%
%
%
%
%
%
%
%

\end{document}